\documentclass[runningheads]{llncs}

\usepackage{eccv}
\usepackage{eccvabbrv}
\usepackage{graphicx}
\usepackage{booktabs}
\usepackage{multirow}
\usepackage[table]{xcolor}
\usepackage{enumitem}
\usepackage{pifont}
\usepackage[most]{tcolorbox}

\newtcblisting{promptblock}{
enhanced,
breakable,
colback=gray!5,
colframe=gray!60,
boxrule=0.4pt,
arc=2pt,
left=6pt,
right=6pt,
top=6pt,
bottom=6pt,
listing only,
listing options={
basicstyle=\ttfamily\small,
breaklines=true,
columns=fullflexible,
keepspaces=true
}
}

\newcommand{\cmark}{%
\textcolor[rgb]{0.10,0.55,0.18}{\ding{51}}%
}

\newcommand{\xmark}{%
\textcolor[rgb]{0.82,0.32,0.32}{\ding{55}}%
}

\definecolor{lightgreen}{HTML}{E0FFE0}
\definecolor{navyblue}{RGB}{58, 117, 196}

\graphicspath{{figures/}}

\usepackage{hyperref}

\makeatletter
\renewcommand{\@fnsymbol}[1]{%
  \ensuremath{\ifcase#1\or *\or\dagger\or\ddagger\or
  \mathsection\or\mathparagraph\or\|\or **\or
  \dagger\dagger\or\ddagger\ddagger\else\@ctrerr\fi}}
\makeatother

\begin{document}

\title{PinpointQA: A Benchmark for Small Object-Centric
Spatial Understanding in Indoor Videos}

\titlerunning{PinpointQA}

\author{
Zhiyu Zhou\inst{1}\thanks{Equal contribution.} \and
Peilin Liu\inst{1}\textsuperscript{\ensuremath{*}} \and
Ruoxuan Zhang\inst{1} \and
Luyang Zhang\inst{1} \and
Cheng Zhang\inst{1} \and
Hongxia Xie\inst{1}\thanks{Corresponding author.} \and
Wen-Huang Cheng\inst{2}
}

\authorrunning{Z.~Zhou et al.}

\institute{
Jilin University, Changchun, China
\and
National Taiwan University, Taipei, Taiwan
}

\maketitle

\begin{abstract}
Reliable embodied interaction in indoor environments requires agents to precisely localize small everyday objects from visual observations. Yet this fundamental capability remains challenging for multimodal large language models (MLLMs), particularly when spatial understanding must be performed from indoor videos. Existing benchmarks study video spatial intelligence and embodied reasoning, but do not directly evaluate whether a model can localize a small target object and express its position with sufficient precision. We introduce PinpointQA, a benchmark built from ScanNet++ and ScanNet200. It contains 1,024 scenes and 10,094 QA pairs across four progressively challenging tasks: Target Presence Verification, Nearest Reference Identification,
Fine-Grained Spatial Description, and Structured Spatial Prediction. Ground-truth annotations are constructed from intermediate spatial representations derived from aligned 3D geometry and instance-level annotations, while evaluated models receive only sampled RGB video frames. Evaluations of representative MLLMs reveal a consistent performance decline across the task progression, with structured spatial prediction remaining particularly challenging, while supervised fine-tuning substantially improves performance. By isolating the spatial grounding capabilities that precede downstream embodied interaction, PinpointQA serves as both a diagnostic benchmark and an effective training resource. The project page is available at
\url{https://rainchowz.github.io/PinpointQA}.

\end{abstract}

\begin{figure}[t]
\centering
\includegraphics[
width=\linewidth,
trim=0 0 0 0,
clip
]{Overview.jpg}
\caption{
Overview of the proposed PinpointQA benchmark.
PinpointQA evaluates small object-centric spatial understanding
in indoor videos through four tasks: Target Presence Verification
(TPV), Nearest Reference Identification (NRI), Fine-Grained
Spatial Description (FSD), and Structured Spatial Prediction
(SSP). The top row shows an example video sequence, the surrounding
panels illustrate the task formats, and the center summarizes the
benchmark statistics and min--max normalized performance of
representative MLLMs.
}
\label{fig:overview}
\end{figure}

\section{Introduction}

In everyday indoor environments, systems that assist users or operate as embodied agents require more than generic video understanding. What matters is whether such a system can reason about a target object:
whether it is present, where it is located, which nearby objects provide
useful cues, and whether this information can be expressed clearly for
downstream use~\cite{grauman2022ego4d,boroushaki2021rfusion}. This need is especially pronounced for small objects such as keys, earbuds,
phones, and chargers. Compared with larger objects, small objects are less salient, more
easily occluded, and more dependent on nearby spatial cues for reliable
localization, making them a distinct challenge for perception and
localization from video. These properties motivate us to study
\emph{small object-centric spatial understanding in video} as a
perceptual and spatial grounding capability that precedes downstream
embodied interaction, rather than as general video
understanding~\cite{feng2025object}.

This problem is also consistent with the basic mechanisms of human visual search. Classical studies show that the visual system does not process all scene
information uniformly, but instead allocates attention preferentially
to regions relevant to the search goal. During search, people rely on an internal target template and narrow the search space through local cues, rather than by fully encoding the entire scene~\cite{desimone1995neural, vickery2005setting, torralba2006contextual, wolfe2021guided}. For small objects, detecting the target is only the
beginning~\cite{he2024decoupling}, since a useful system must also
identify nearby reference objects and organize their relations into
precise spatial descriptions~\cite{huang2024chat,liang2025referdino}. We therefore model the problem as a progressive capability chain that moves from target perception to reference grounding, fine-grained spatial description, and structured spatial prediction.

Recent progress in egocentric and long-video datasets, together with large-scale 3D scene resources with dense geometric and instance-level annotations, has made small object-centric spatial understanding increasingly feasible to study and benchmark~\cite{dai2017scannet, yeshwanth2023scannet++, rozenberszki2022language, grauman2022ego4d, damen2022rescaling, mangalam2023egoschema}. Meanwhile, the rapid improvement of multimodal large models has made it increasingly important to systematically evaluate their capabilities on this problem~\cite{team2023gemini, singh2025openai, bai2025qwen3, an2025llava, wang2025internvl3}. Recent benchmarks have advanced the evaluation of video spatial intelligence, multi-hop spatial reasoning, precise spatial-temporal understanding, and embodied spatial understanding~\cite{yang2025thinking, zhu2026video, li2025sti, du2024embspatial, lin2025ostbench, yuan2025videorefer, wang2025site}. Small object-centric and diagnostic benchmarks have further highlighted the difficulty of tiny targets under occlusion, weak salience, and strong reliance on local context~\cite{patraucean2023perception, zhu2023egoobjects, shen2024learning}. However, these efforts leave an important gap between video perception
and embodied reasoning: they neither directly evaluate whether a model
can localize a small target object from video observations and recover
its supporting surface, nearby references, spatial relations, and
metric distances, nor explicitly diagnose the progression from target
perception to reference grounding, precise spatial description, and
structured spatial output.

To fill this gap, we introduce PinpointQA, the first dedicated benchmark for small object-centric spatial understanding in indoor videos. Built from ScanNet++~\cite{yeshwanth2023scannet++} and ScanNet200~\cite{rozenberszki2022language}, where the latter extends the original ScanNet v2~\cite{dai2017scannet} scene data and splits with a finer-grained label space, PinpointQA contains 1,024 scenes and 10,094 QA pairs, and supports both systematic evaluation and task-oriented fine-tuning. As shown in Figure~\ref{fig:overview}, PinpointQA is organized around
four tasks, Target Presence Verification (TPV), Nearest Reference
Identification (NRI), Fine-Grained Spatial Description (FSD), and
Structured Spatial Prediction (SSP), while the representative examples,
benchmark statistics, and baseline results provide an overall view of
the proposed benchmark.

Our main contributions are as follows:
\begin{itemize}
    \item \textbf{A dedicated benchmark.}
    We present the first dedicated benchmark for small
    object-centric spatial understanding in indoor videos, targeting a
    perceptual and spatial grounding capability that supports embodied
    reasoning in physical environments.

    \item \textbf{A progressive task suite.}
    We formulate four benchmark tasks, which define a progressive capability chain from target presence verification to reference grounding, fine-grained spatial description, and
    structured spatial prediction.

    \item \textbf{Construction and evaluation grounded in geometry.}
We construct reproducible annotations from ScanNet++ and ScanNet200
using aligned geometry, while models are evaluated using only video
input. Through comprehensive evaluation and fine-tuning on two
representative models, we further show that PinpointQA serves as both a
diagnostic benchmark and a useful supervision resource.
\end{itemize}

\section{Related Work}

\begin{table}[t]
\centering
\caption{Comparison with related benchmarks. From left to right, the columns denote Benchmark (\textit{Bench.}), Spatial Reasoning (\textit{Spatial}), Continuous Scene Observation (\textit{Obs.}), Everyday Indoor Object Context (\textit{Indoor}), and Small Object Localization (\textit{Local.}).}
\label{tab:benchmark_comparison}
\setlength{\tabcolsep}{2.5pt}
\renewcommand{\arraystretch}{1.05}
\begin{tabular*}{\columnwidth}{@{\extracolsep{\fill}}lcccc@{}}
\toprule
\textbf{Bench.} & \textbf{Spatial} & \textbf{Obs.} & \textbf{Indoor} & \textbf{Local.} \\
\midrule
Open3D-VQA~\cite{zhang2025open3d}       & \cmark & \xmark & \xmark & \xmark \\
EmbSpatial-Bench~\cite{du2024embspatial} & \cmark & \xmark & \cmark & \xmark \\
V-STaR~\cite{cheng2025v}           & \cmark & \cmark & \xmark & \xmark \\
Video-MSR~\cite{zhu2026video}        & \cmark & \cmark & \xmark & \xmark \\
STI-Bench~\cite{li2025sti}        & \cmark & \cmark & \cmark & \xmark \\
VSI-Bench~\cite{yang2025thinking}        & \cmark & \cmark & \cmark & \xmark \\
\midrule
\textbf{PinpointQA} & \textbf{\cmark} & \textbf{\cmark} & \textbf{\cmark} & \textbf{\cmark} \\
\bottomrule
\end{tabular*}
\end{table}

\subsection{Spatial Understanding Benchmarks}
Recent benchmarks have explored spatial understanding in videos and
embodied 3D scenes from multiple perspectives. VSI-Bench
~\cite{yang2025thinking} studies visual-spatial intelligence from
sequential observations, while V-STaR~\cite{cheng2025v} and
STI-Bench~\cite{li2025sti} focus on spatio-temporal reasoning and
precise spatial-temporal understanding. Video-MSR~\cite{zhu2026video}
further examines multi-hop spatial reasoning in dynamic videos, whereas
Open3D-VQA~\cite{zhang2025open3d} and
EmbSpatial-Bench~\cite{du2024embspatial} study spatial relation
reasoning and embodied spatial understanding for MLLMs. However, these benchmarks mainly focus on general spatial reasoning or
broader scene and embodied understanding, rather than a progressive
task chain centered on small object localization, even though reliable
target grounding provides an important perceptual and spatial
foundation for downstream embodied behavior. Table
~\ref{tab:benchmark_comparison} compares representative benchmarks
along four dimensions and shows that existing work generally covers
only part of spatial reasoning, continuous scene observation, indoor
object context, and small object localization. PinpointQA addresses this
gap by integrating these abilities within a progressive diagnostic
framework centered on the localization of small objects.

\subsection{Small Object Perception and Understanding Benchmarks}
Research on small object understanding has highlighted the challenges
posed by weak salience, occlusion, and limited visual evidence.
EgoObjects~\cite{zhu2023egoobjects} studies detailed small object understanding in egocentric videos, while Perception Test~\cite{patraucean2023perception} includes diagnostic tasks involving
small object permanence, occlusion, and containment. Although these
benchmarks move beyond generic video understanding and directly examine
the difficulty of small targets, they mainly emphasize object
understanding or diagnostic video reasoning and do not evaluate whether
a model can localize a target object and express its position with
sufficient precision for downstream applications.

\section{PinpointQA Benchmark}

To systematically evaluate small object-centric spatial understanding in
videos, we introduce PinpointQA, a benchmark designed to assess whether
MLLMs can detect a target object, localize it using surrounding visual
cues, and express its position in forms useful to both humans and
downstream systems. These abilities provide the perceptual and spatial
grounding required by embodied tasks that depend on reliable object
localization. Figure~\ref{fig:overview} presents the four tasks and the
overall structure of PinpointQA, while the following sections describe
the task formulation, construction pipeline, and benchmark statistics.

\subsection{Task Formulation}
\label{subsec:task_formulation}

Small object-centric spatial understanding is not a single-step ability. A model must first determine whether the target object is present, then anchor it with nearby reference objects, further describe its location in a precise and human-readable way, and finally organize the same grounded spatial information into a structured form for downstream use. To capture this progression, we define four tasks in PinpointQA: \textit{Target Presence Verification (TPV)}, \textit{Nearest Reference Identification (NRI)}, \textit{Fine-Grained Spatial Description (FSD)}, and \textit{Structured Spatial Prediction (SSP)}. Together, these tasks form a progressive capability chain from target presence verification to reference-based grounding, then to natural-language spatial description, and ultimately to structured spatial output.

\textbf{Target Presence Verification (TPV).} TPV isolates the entry-level requirement of small object-centric spatial understanding: determining whether a target object appears in the video at all. Unlike later tasks that require localization or spatial expression, TPV focuses only on presence verification under realistic indoor conditions, where small objects are often difficult to notice because of clutter, occlusion, and viewpoint changes. The task therefore asks the model to judge whether the target is present in the observed scene, establishing the basis on which all later localization-oriented reasoning depends.

\textbf{Nearest Reference Identification (NRI).} NRI examines whether the model can move beyond target presence and begin to ground the target in its local spatial context. Rather than simply recognizing nearby objects, the key challenge is to determine which reference object provides the nearest cue for localization. Concretely, the model is asked to identify the reference object closest to the target, excluding the supporting surface. Successful performance on this task requires the model to capture local spatial proximity around the target, making NRI the first step from target presence to local spatial grounding.

\textbf{Fine-Grained Spatial Description (FSD).} FSD examines whether the model can express the target location as a natural-language spatial description based on local spatial cues. Unlike TPV and NRI, which remain closed-form tasks, FSD requires the model to organize multiple pieces of spatial information into coherent free-form text. A correct answer should bring together the supporting surface, nearby reference objects, relative directions, and distances, so that the target location can be conveyed clearly and precisely to a human reader~\cite{kato2023arkitscenerefer}. This task therefore evaluates not only whether the model identifies the relevant local spatial cues, but also whether it can express them as a fine-grained spatial description.

\textbf{Structured Spatial Prediction (SSP).} SSP examines whether the model can further organize the target location into a structured spatial output. Unlike FSD, which expresses spatial information in free-form natural language, SSP requires the model to preserve the same localization-relevant spatial information in a more standardized format. The expected output includes the key fields that directly support localization, such as the supporting surface, nearby reference objects, and their corresponding distances. As the final task in the chain, SSP tests whether the model can not only capture local spatial information, but also present it in a form that can be more easily parsed and used by downstream systems.

\subsection{Dataset Construction}
\label{subsec:dataset_construction}

As illustrated in Figure~\ref{fig:data_pipeline}, PinpointQA is constructed in four stages: data collection, scene curation, task-specific QA generation, and quality control. Rather than generating QA pairs directly from the source annotations, we first derive an intermediate spatial representation for each valid target from the aligned 3D meshes and instance-level annotations, and then instantiate these representations as QA pairs for the four tasks defined in Section~\ref{subsec:task_formulation}. 

\textbf{Data Collection.}
PinpointQA is built on the ScanNet++ and ScanNet200 datasets, which
provide 3D reconstructions and semantic annotations of real indoor
scenes. Both datasets consist of reconstructions of captured indoor
environments rather than synthetically generated scenes. ScanNet++ provides detailed and accurately aligned scene geometry, while ScanNet200 extends the ScanNet v2 scenes with a more detailed label space containing 200 semantic categories. For benchmark construction, we use the aligned 3D meshes and annotations
of individual object instances to identify candidate objects and
construct their intermediate spatial representations during scene
curation.

\begin{figure*}[t]
    \centering
    \includegraphics[width=\textwidth, trim=0 1 1 1, clip]{figures/Data-Pineline.jpg}
    \caption{
Construction pipeline of PinpointQA. Starting from the aligned 3D meshes and instance-level annotations of ScanNet++ and ScanNet200, we select candidate small object targets, construct intermediate spatial
representations through scene curation, and instantiate them as TPV, NRI, FSD, and SSP QA pairs. The pipeline combines rule-based filtering with
iterative manual review for quality control.
    }
    \label{fig:data_pipeline}
\end{figure*}

\textbf{Scene Curation.} Given the aligned meshes and object annotations collected in the
previous stage, scene curation identifies valid small object targets and constructs an intermediate spatial representation for each target. We define small objects by semantic category rather than by their projected image area, since apparent size varies with viewpoint and viewing distance. Our vocabulary covers 102 categories of portable or movable indoor
objects, particularly objects that can be held in hand or are commonly
placed on tables. Equivalent source labels, such as \textit{phone},
\textit{mobile phone}, and \textit{smartphone}, are counted as the same
category, while their original wording is retained in the generated QA
pairs. The complete vocabulary and the grouping of equivalent labels
are provided in Appendix~A. After converting the source labels to lowercase, we select target
candidates by exact matching against the vocabulary and exclude
furniture, architectural elements, and objects at the scale of a room.
We discard scenes without a valid target and remove any target whose
normalized source label corresponds to more than one eligible object in
the same scene, thereby avoiding ambiguous references.

For each retained target, we form a local neighborhood from the annotated objects whose mesh centroids lie within 1.0~m of the target centroid, excluding structural categories. Each object centroid is computed as the mean of its assigned mesh vertices. The centroid distance is used only to determine which objects belong to the local neighborhood. For each target and neighboring object, we then compute the shortest distance between their mesh vertices, their horizontal separation in the XY plane, and their vertical offset along the Z axis. The shortest mesh distance is used as the metric distance in the generated annotations. A difference in centroid height greater than 5~cm determines whether the target is above or below the neighboring
object; smaller differences are treated as the same level. Together, the object labels and these geometric measurements form the intermediate spatial representation used in the next stage to determine supporting surfaces, select reference objects, derive semantic relations, and
generate the four QA tasks.

\textbf{Task-specific QA Generation.} We use the intermediate spatial representations constructed during scene curation to generate QA pairs for the four tasks. We first identify candidate supporting surfaces and reference objects from the neighboring objects using their semantic labels and geometric measurements relative to the target. Supporting surfaces are determined by jointly considering object category, relative height, horizontal separation, and the shortest distance between mesh vertices, with direct contact and geometrically consistent placement receiving the highest priority. A shortest mesh distance of no more than 2~cm is treated as contact. When multiple candidates are available, semantic suitability and geometric proximity determine the selected supporting surface. For FSD and SSP, when multiple reference instances share the same label, we retain the instance with the shortest mesh distance to the target. The geometric measurements are then mapped to nine semantic relations: \textit{on}, \textit{under}, \textit{above},
\textit{below}, \textit{side\_above}, \textit{side\_below}, \textit{next\_to}, \textit{near}, and \textit{attached\_to}. This mapping considers object category, contact, relative height, horizontal separation, and whether the target and reference share the same supporting surface.

For TPV, positive queries use retained target labels, while negative queries are sampled from the remaining labels in the target vocabulary. For NRI, the correct option is the valid reference with the shortest mesh distance to the target after excluding the supporting surface, floor and wall categories, and objects in direct contact. Each NRI question contains three distinct distractors, and cases with a near tie between the closest reference labels are discarded. For FSD, the selected supporting surface is used as the main anchor when available; otherwise, the closest valid reference becomes the anchor. Up to two additional references are included with their semantic relations and distances rounded to the nearest centimeter. For SSP, the target, the selected supporting surface, and up to two
closest references are encoded in a parsable JSON object, with
distances retained to one decimal place. The four tasks thus share the same underlying intermediate spatial representations while applying different selection and output constraints.

\textbf{Quality Control.} Our quality control combines automatic filtering based on predefined rules with iterative manual review. At the scene and object levels, the automatic filters remove invalid
labels, structural categories, and target instances whose labels are
ambiguous within a scene. At the QA level, TPV answers are restricted
to \textit{Yes} or \textit{No}; NRI questions must contain four distinct
options and a uniquely determined nearest valid reference; FSD must
provide a complete natural language answer grounded in at least one
valid spatial anchor; and SSP must be directly parsable as a JSON object
containing the required \texttt{target}, \texttt{support\_surface}, and
\texttt{references} fields. Any sample or scene that fails these
requirements is excluded. In each manual review round, we inspect approximately 100 QA pairs from 10--15 scenes. The review examines whether NRI excludes the supporting surface from its reference candidates, whether FSD preserves the
intended supporting surface and nearby references, whether distances and semantic relations remain consistent with the underlying geometric measurements, whether the question and answer templates are natural and well-formed, and whether the resulting localization is plausible to a human observer. Problems identified during a review round are used to revise the QA generation rules, filtering criteria, and templates before the next round, while the underlying intermediate spatial representations remain unchanged.

\definecolor{tablegreen}{HTML}{EEF5FF}

\subsection{Benchmark Statistics}

\begin{figure}[t]
    \centering
    \includegraphics[width=1\columnwidth, trim=0 0 0 0, clip]{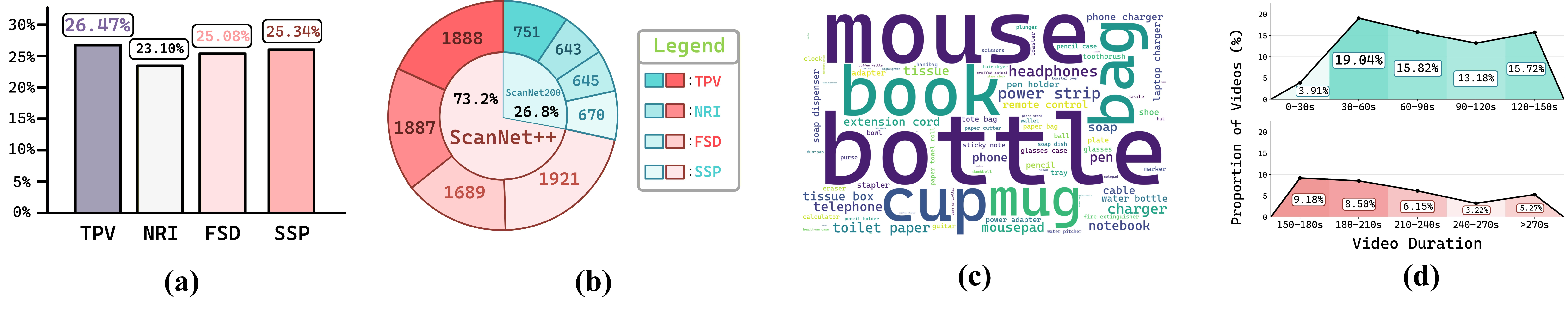}
    \caption{Benchmark statistics: (a) task distribution, (b) composition
by data source and counts for each task, (c) categories of target
objects shown as a word cloud, and (d) video duration distribution.}
    \label{fig:stats}
\end{figure}

Figure~\ref{fig:stats} summarizes PinpointQA in terms of task
distribution, data source composition, target categories, and video
duration. The benchmark contains 1,024 scenes and 10,094 QA pairs. TPV
and NRI contain 2,672 and 2,332 QA pairs, accounting for 26.47\% and
23.10\% of the benchmark, respectively, while FSD and SSP contain 2,532
and 2,558 QA pairs, accounting for 25.08\% and 25.34\%. The difference
between the largest and smallest task proportions is only 3.37
percentage points, indicating that no individual task dominates the
benchmark.

ScanNet++ contributes 7,385 QA pairs, accounting for 73.2\% of the
benchmark, while ScanNet200 contributes the remaining 2,709 QA pairs,
or 26.8\%. Both data sources provide samples for all four tasks. The
target vocabulary covers 102 categories and includes a broad range of
everyday small objects.

The video duration distribution contains both short and long
observations. The interval from 30 to 60~s is the largest, containing
195 videos and accounting for 19.04\% of all videos. Overall, 693 videos are no longer than 150~s, accounting for 67.68\%
of all videos, while 331 videos exceed 150~s, accounting for 32.32\%.
Among the latter, 54 videos exceed 270~s, corresponding to 5.27\% of
the full benchmark.

\section{Experiments}

\subsection{Experimental Setup}

\textbf{Evaluated Models.}
We evaluate three groups of MLLMs on the proposed benchmark. The
proprietary models are GPT-5.4~\cite{singh2025openai} and Kimi
K2.5~\cite{team2026kimi}. The open-source models are LLaVA-OneVision-1.5-8B~\cite{an2025llava},
Qwen3-VL-8B-Instruct~\cite{bai2025qwen3}, and
InternVL3.5-8B-Instruct~\cite{wang2025internvl3}. We additionally
evaluate Spatial-MLLM-v1.1-Instruct-820K~\cite{wu2025spatial},
SenseNova-SI-1.3-InternVL3-8B~\cite{cai2025scaling}, and
Cambrian-S-7B~\cite{yang2025cambrian}. To examine the utility of
PinpointQA as a training resource, we perform LoRA fine-tuning on two
representative backbones, yielding Qwen3-VL-8B-Instruct-SFT and
InternVL3.5-8B-Instruct-SFT~\cite{hu2022lora}.

\textbf{Benchmark Split.}
We partition the benchmark by scene into training, validation, and
evaluation splits that approximately follow a 6:2:2 ratio and contain
6,121, 1,954, and 2,019 samples, respectively. During partitioning, we
jointly preserve the proportions of the two data sources and the
distribution of the four task types. Using the same scene partition for
all four tasks prevents data leakage across the splits.

\textbf{Inference and Fine-tuning.}
For inference, we uniformly sample 64 RGB frames from each input video,
and these frames constitute the sole visual input to every evaluated
model. Each model additionally receives the question and a task
instruction specifying the required answer format, applicable rules,
and examples when needed. No depth information, camera poses,
reconstructed geometry, or 3D annotations are provided, so models must
estimate spatial relations and distances solely from visual cues in the
sampled RGB frames and their learned spatial reasoning capabilities.

The reconstructed geometry and 3D annotations described in
Section~\ref{subsec:dataset_construction} are used only to construct the
ground truth. The resulting protocol uses video observations as the only visual
input, while the annotations derived from geometry provide an accurate,
physically grounded, and reproducible basis for evaluation. Across models, we keep the task instructions and question templates
consistent, make only minor adaptations for different model interfaces,
and use greedy decoding for generation.

For fine-tuning, we train on the training split, select the best checkpoint using the validation split, and run LoRA fine-tuning for two epochs. This controlled setting examines whether PinpointQA provides effective supervision for small object-centric spatial understanding, without introducing separate tuning strategies for individual tasks.

\subsection{Evaluation Metrics}

Because TPV, NRI, FSD, and SSP have different output formats, we evaluate them using task-specific metrics and normalize all task scores to the interval from 0 to 1.

\textbf{Closed-ended Tasks.}
TPV and NRI have finite answer spaces with a single correct answer, so we use exact-match accuracy. For TPV, we extract the predicted \textit{Yes} or \textit{No} label and compare it with the ground truth. For NRI, we extract the predicted option letter and compare it with the correct choice. An output from which no valid label or option can be extracted receives a score of zero.

\textbf{Fine-Grained Spatial Description.}
Lexical metrics and embedding metrics, including BLEU, ROUGE, and
BERTScore~\cite{papineni2002bleu,lin2004rouge,zhang2019bertscore},
as well as exact string matching, do not directly evaluate whether a
description preserves the correct grounded spatial information. We
therefore use GPT-5.4 as an LLM judge~\cite{liu2023g} to compare each
prediction with the ground truth using a rubric with a maximum score of
10.

The rubric assigns up to 3 points to the main location and supporting
surface, up to 2 points each to key reference objects, spatial
relations, and numeric distances, and 1 point to clarity. Consequently,
9 of the 10 points measure grounded spatial content, ensuring that
language clarity cannot compensate for incorrect localization. The
judge penalizes unsupported reference objects and descriptions that
replace a directional relation with vague proximity, while numeric
distances are evaluated using tolerances that increase with the ground
truth distance. We sum the scores for the five dimensions and divide
the result by 10 to obtain the normalized FSD score.

\textbf{Structured Spatial Prediction.}
Strict JSON exact matching is unsuitable for SSP because field order,
minor formatting differences, and equivalent label forms should not
automatically produce a score of zero. The evaluator first attempts to recover a JSON object, applies
conservative normalization to object labels, supporting surface labels,
and relation tokens, and verifies its basic structure. If no JSON object
can be recovered, the prediction receives a score of zero. Since the
queried target is already specified in the question, SSP scoring focuses
on the localization fields. The predicted references are aligned with the ground truth reference slots before scoring supporting surface, reference object identity, spatial relation, and distance.

The supporting surface and reference object scores require exact matches after normalization. Spatial relations receive full credit for an exact match and half credit for the limited semantically close pairs \textit{next\_to} and \textit{near}, \textit{under} and \textit{below}, or \textit{on} and \textit{attached\_to}. Relation credit is assigned only after the reference object is matched, and distance credit additionally requires a compatible relation. Distance errors are evaluated using tolerances that increase with the ground truth distance. The final SSP score is

\[
S_{\mathrm{SSP}}
= 0.4 S_{\mathrm{surface}}
+ 0.2 S_{\mathrm{object}}
+ 0.2 S_{\mathrm{relation}}
+ 0.2 S_{\mathrm{distance}}.
\]

The supporting surface receives the largest individual weight because
it provides the primary localization anchor, while the supporting
surface, reference object identity, and spatial relation together
account for 0.8 of the final score. Distance serves as a more precise
geometric constraint after the categorical spatial structure has been
correctly grounded. These weights express the intended emphasis of
PinpointQA rather than a uniquely optimal metric, and alternative
weighting choices may affect relative model rankings.

\textbf{Aggregation.}
For each task, Micro is the mean over all QA pairs, whereas Macro first
computes a mean within each scene and then assigns equal weight to the
result from every scene. Avg Micro and Avg Macro are the arithmetic
means of the corresponding scores across TPV, NRI, FSD, and SSP.
Detailed scoring rules and implementation settings are provided in
Appendix~B.

\subsection{Main Results on the Proposed Benchmark}
\label{sec:main_results}

Table~\ref{tab:main_results} summarizes the performance of all evaluated
models across the four tasks, while Figure~\ref{fig:overview} compares
representative models using a radar chart. We analyze these results in
terms of overall competitiveness, progressive capability breakdown,
failure profiles across stages, and training utility.

\textbf{Overall competitiveness.}
Qwen3-VL-8B-Instruct-SFT achieves the best overall performance, reaching 0.48 in Avg Micro and 0.49 in Avg Macro, followed by InternVL3.5-8B-Instruct-SFT with 0.45 in both metrics. Among the proprietary models, Kimi K2.5 outperforms GPT-5.4, achieving 0.42 in Avg Micro and 0.44 in Avg Macro, compared with 0.38 and 0.40 for GPT-5.4. Among the open-source models without additional tuning,
Qwen3-VL-8B-Instruct performs best, reaching 0.39 in Avg Micro and 0.40
in Avg Macro, thereby exceeding GPT-5.4 in Avg Micro and matching it in
Avg Macro. More broadly, the rankings under Micro and Macro are highly consistent, suggesting that the observed differences are not driven primarily by scenes that contribute larger numbers of QA pairs.

\textbf{Progressive capability breakdown.}
Scores decrease monotonically from TPV to SSP for almost all models. For example, Qwen3-VL-8B-Instruct-SFT decreases from 0.83 in Micro and 0.84 in Macro on TPV to 0.29 in both metrics on SSP. Kimi K2.5 decreases from 0.80 and 0.84 to 0.15 in both metrics, while Qwen3-VL-8B-Instruct decreases from 0.78 and 0.80 to 0.12 in both metrics. The four tasks impose progressively more demanding requirements: TPV
asks whether the target is visible, NRI additionally requires the
identification of an appropriate reference object, FSD requires a
description that correctly integrates the supporting surface,
reference objects, spatial relations, and distances, and SSP requires
all these components to remain mutually consistent in a structured
output. Consequently, strong target presence verification does not necessarily translate into reliable reference grounding or structured spatial
prediction. This progressive decline supports the four-task design by separating target presence verification from increasingly demanding spatial reasoning and output organization. This distinction is particularly relevant to embodied systems because
recognizing a target is only an initial step, while reliable downstream
use also requires the model to recover where the target is supported
and how it relates to nearby spatial anchors.

\begin{table}[t]
\centering
\caption{Main results on PinpointQA across the four tasks. For each task, Micro and Macro report the mean over all its QA pairs and the mean of per-scene scores, respectively. Avg is the arithmetic mean across the four tasks. Bold denotes the best performance in each column.}
\label{tab:main_results}

\resizebox{\linewidth}{!}{%
\begin{tabular}{l cc cc cc cc cc}

\toprule
\textbf{Model} & \multicolumn{2}{c}{\textbf{TPV}} & \multicolumn{2}{c}{\textbf{NRI}} & \multicolumn{2}{c}{\textbf{FSD}} & \multicolumn{2}{c}{\textbf{SSP}} & \multicolumn{2}{c}{\textbf{Avg}} \\
\cmidrule(r){2-3} \cmidrule(r){4-5} \cmidrule(r){6-7} \cmidrule(r){8-9} \cmidrule(r){10-11}
 & Micro & Macro & Micro & Macro & Micro & Macro & Micro & Macro & Micro & Macro \\
\midrule
\rowcolor{tablegreen} \multicolumn{11}{l}{\textit{Proprietary Models}} \\
GPT-5.4~\cite{singh2025openai} & 0.65 & 0.69 & 0.39 & 0.42 & 0.31 & 0.32 & 0.15 & 0.16 & 0.38 & 0.40 \\
Kimi K2.5~\cite{team2026kimi} & 0.80 & \textbf{0.84} & 0.42 & 0.44 & 0.32 & 0.33 & 0.15 & 0.15 & 0.42 & 0.44 \\
\midrule
\rowcolor{tablegreen} \multicolumn{11}{l}{\textit{Open-source Models}} \\
LLaVA-OneVision-1.5-8B~\cite{an2025llava} & 0.76 & 0.79 & 0.30 & 0.30 & 0.26 & 0.27 & 0.07 & 0.06 & 0.35 & 0.36 \\
Qwen3-VL-8B-Instruct~\cite{bai2025qwen3} & 0.78 & 0.80 & 0.37 & 0.37 & 0.28 & 0.29 & 0.12 & 0.12 & 0.39 & 0.40 \\
InternVL3.5-8B-Instruct~\cite{wang2025internvl3} & 0.65 & 0.70 & 0.36 & 0.38 & 0.25 & 0.26 & 0.09 & 0.10 & 0.34 & 0.36 \\
Spatial-MLLM-v1.1-Instruct-820K~\cite{wu2025spatial} & 0.52 & 0.51 & 0.30 & 0.30 & 0.21 & 0.20 & 0.00 & 0.00 & 0.26 & 0.25 \\
SenseNova-SI-1.3-InternVL3-8B~\cite{cai2025scaling} & 0.64 & 0.66 & 0.36 & 0.40 & 0.15 & 0.16 & 0.12 & 0.13 & 0.32 & 0.34 \\
Cambrian-S-7B~\cite{yang2025cambrian} & 0.73 & 0.78 & 0.33 & 0.35 & 0.24 & 0.25 & 0.05 & 0.06 & 0.34 & 0.36 \\
\midrule
\rowcolor{tablegreen} \multicolumn{11}{l}{\textit{Fine-tuned Models}} \\
\textbf{Qwen3-VL-8B-Instruct-SFT} & \textbf{0.83} & \textbf{0.84} & \textbf{0.44} & \textbf{0.45} & \textbf{0.36} & \textbf{0.37} & \textbf{0.29} & \textbf{0.29} & \textbf{0.48} & \textbf{0.49} \\
InternVL3.5-8B-Instruct-SFT & 0.82 & 0.82 & 0.41 & 0.39 & 0.34 & 0.36 & 0.23 & 0.24 & 0.45 & 0.45 \\
\bottomrule

\end{tabular}%
}

\end{table}

\textbf{Failure profiles across stages.}
The task-level results reveal distinct performance bottlenecks across models. LLaVA-OneVision-1.5-8B achieves 0.76 in Micro and 0.79 in Macro on TPV, but decreases to 0.26 and 0.27 on FSD and 0.07 and 0.06 on SSP. SenseNova-SI-1.3-InternVL3-8B attains 0.36 and 0.40 on NRI but only 0.15 and 0.16 on FSD, showing substantial deterioration between reference identification and natural-language spatial description. Spatially oriented models also remain limited on SSP:
Spatial-MLLM-v1.1-Instruct-820K obtains 0.00 in both SSP metrics, while
Cambrian-S-7B achieves 0.73 and 0.78 on TPV but only 0.05 and 0.06 on
SSP. Together, these results suggest that general spatial reasoning does not
automatically transfer to structured spatial prediction for small
objects from video observations, since reasonable target presence verification can
coexist with unreliable supporting surface prediction, reference
grounding, relation estimation, and distance estimation. By reporting
results for the individual tasks, PinpointQA reveals where performance
deteriorates within the capability hierarchy, which would be obscured
by a single aggregated score.

Qualitative inspection reveals failures at multiple stages: errors in
target grounding and nearest reference identification can occur before
structured prediction, while outputs that satisfy the required JSON
format may still contain supporting surface mismatches, reference
hallucination or drift, relation inconsistencies, inaccurate distances,
or cross-field conflicts. Representative cases covering these failure
types are provided in Appendix~C.

\textbf{Dataset as a training resource.}
Fine-tuning produces consistent improvements for both evaluated model families. Qwen3-VL-8B-Instruct improves from 0.39 and 0.40 in Avg Micro and Avg Macro to 0.48 and 0.49, respectively. InternVL3.5-8B-Instruct improves from 0.34 and 0.36 to 0.45 in both metrics. The largest gains occur on SSP, where Qwen improves from 0.12 in both
metrics to 0.29 in both metrics, while InternVL improves from 0.09 and
0.10 to 0.23 and 0.24. These improvements demonstrate that PinpointQA can serve not only as a diagnostic benchmark but also as a task-oriented supervision resource. Nevertheless, SSP performance remains substantially lower than TPV, NRI, and FSD even after fine-tuning, indicating that small object-centric structured spatial prediction remains the primary bottleneck for current models.

\subsection{Human Localization with FSD Guidance}
To examine whether grounded spatial reasoning can support human
localization, we test whether FSD descriptions help people locate small
objects in an auxiliary online evaluation with three settings.
Participants receive only the question and video frames in the
Unguided setting, additionally receive the FSD description generated by
Qwen3-VL-8B-Instruct-SFT in the Model-Assisted setting, and receive the
corresponding ground truth description in the GT-Assisted setting. The
evaluation includes 60 FSD samples, with 20 assigned to each setting,
and reports the mean results across five participants. Further details
of the protocol and interface are provided in Appendix~D.

\begin{figure}[tbp]
    \centering
    \includegraphics[
        width=\linewidth,
        trim=3 45 5 15,
        clip
    ]{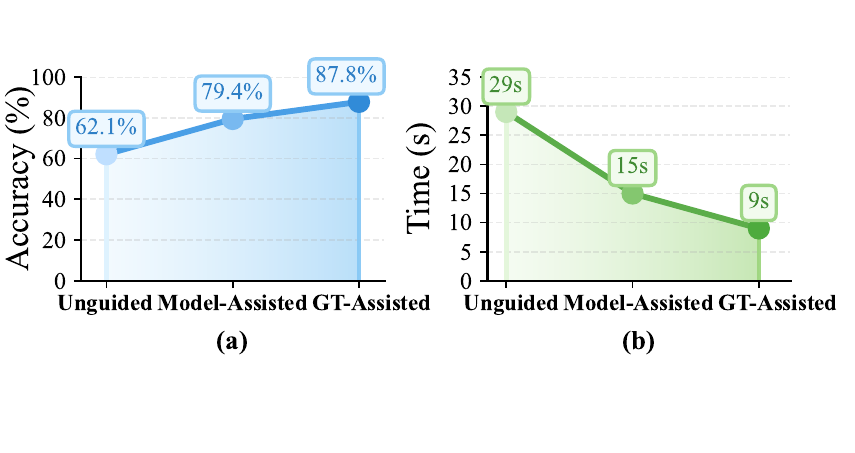}
    \caption{
Human localization accuracy and average completion time under
Unguided, Model-Assisted, and GT-Assisted settings.
}
    \label{fig:human_study}
\end{figure}

Participants browse the available video frames, select the frame in which
the target appears, and click its estimated location. Completion time is
measured from when the frames are revealed until submission, covering both
frame selection and spatial localization. Accuracy is evaluated against
manually annotated target points. An incorrect frame receives a score of
zero; otherwise, a soft score is assigned based on the normalized distance
between the selected and annotated locations. This evaluation therefore
reflects both correct-frame identification and localization precision.

As shown in Figure~\ref{fig:human_study}, both guided settings outperform Unguided: Model-Assisted and
GT-Assisted improve accuracy from 62.1\% to 79.4\% and 87.8\%,
respectively, while reducing completion time from 29 s to 15 s and
9 s. The remaining gap indicates that FSD utility depends on
description accuracy; we therefore treat this study as auxiliary
evidence rather than a deployment evaluation.
\section{Conclusion}

We introduced PinpointQA, a benchmark for small object-centric spatial
understanding in indoor videos, with 1,024 ScanNet++ and ScanNet200 scenes
and 10,094 QA pairs across TPV, NRI, FSD, and SSP. Using video-only input
and geometry-derived annotations, it evaluates target verification, reference
grounding, spatial description, and structured prediction. Experiments reveal
a consistent decline across this progression, particularly on SSP, while
fine-tuning substantially improves performance. PinpointQA thus serves as
both a diagnostic benchmark and a supervision resource for spatial grounding
in embodied systems, with potential extensions to interactive navigation and
manipulation.

\section*{Acknowledgments}
This work was funded by the National Natural Science Foundation of China (Grant No.~62406126).

\bibliographystyle{splncs04}
\bibliography{ref}

@String(ICLR  = {Int. Conf. Learn. Represent.})

@String(ICLR  = {ICLR})

@inproceedings{grauman2022ego4d,
  title={Ego4d: Around the world in 3,000 hours of egocentric video},
  author={Grauman, Kristen and Westbury, Andrew and Byrne, Eugene and Chavis, Zachary and Furnari, Antonino and Girdhar, Rohit and Hamburger, Jackson and Jiang, Hao and Liu, Miao and Liu, Xingyu and others},
  booktitle={Proceedings of the IEEE/CVF conference on computer vision and pattern recognition},
  pages={18995--19012},
  year={2022}
}

@inproceedings{boroushaki2021rfusion,
  title={Rfusion: Robotic grasping via rf-visual sensing and learning},
  author={Boroushaki, Tara and Perper, Isaac and Nachin, Mergen and Rodriguez, Alberto and Adib, Fadel},
  booktitle={Proceedings of the 19th ACM conference on embedded networked sensor systems},
  pages={192--205},
  year={2021}
}

@inproceedings{zhu2023egoobjects,
  title={Egoobjects: A large-scale egocentric dataset for fine-grained object understanding},
  author={Zhu, Chenchen and Xiao, Fanyi and Alvarado, Andr{\'e}s and Babaei, Yasmine and Hu, Jiabo and El-Mohri, Hichem and Culatana, Sean and Sumbaly, Roshan and Yan, Zhicheng},
  booktitle={Proceedings of the IEEE/CVF international conference on computer vision},
  pages={20110--20120},
  year={2023}
}

@article{patraucean2023perception,
  title={Perception test: A diagnostic benchmark for multimodal video models},
  author={Patraucean, Viorica and Smaira, Lucas and Gupta, Ankush and Recasens, Adria and Markeeva, Larisa and Banarse, Dylan and Koppula, Skanda and Malinowski, Mateusz and Yang, Yi and Doersch, Carl and others},
  journal={Advances in Neural Information Processing Systems},
  volume={36},
  pages={42748--42761},
  year={2023}
}

@article{desimone1995neural,
  title={Neural mechanisms of selective visual attention},
  author={Desimone, Robert and Duncan, John and others},
  journal={Annual review of neuroscience},
  volume={18},
  number={1},
  pages={193--222},
  year={1995}
}

@article{vickery2005setting,
  title={Setting up the target template in visual search},
  author={Vickery, Timothy J and King, Li-Wei and Jiang, Yuhong},
  journal={Journal of vision},
  volume={5},
  number={1},
  pages={8--8},
  year={2005},
  publisher={The Association for Research in Vision and Ophthalmology}
}

@article{torralba2006contextual,
  title={Contextual guidance of eye movements and attention in real-world scenes: the role of global features in object search.},
  author={Torralba, Antonio and Oliva, Aude and Castelhano, Monica S and Henderson, John M},
  journal={Psychological review},
  volume={113},
  number={4},
  pages={766},
  year={2006},
  publisher={American Psychological Association}
}

@article{wolfe2021guided,
  title={Guided Search 6.0: An updated model of visual search},
  author={Wolfe, Jeremy M},
  journal={Psychonomic bulletin \& review},
  volume={28},
  number={4},
  pages={1060--1092},
  year={2021},
  publisher={Springer}
}

@article{wang2025internvl3,
  title={Internvl3. 5: Advancing open-source multimodal models in versatility, reasoning, and efficiency},
  author={Wang, Weiyun and Gao, Zhangwei and Gu, Lixin and Pu, Hengjun and Cui, Long and Wei, Xingguang and Liu, Zhaoyang and Jing, Linglin and Ye, Shenglong and Shao, Jie and others},
  journal={arXiv preprint arXiv:2508.18265},
  year={2025}
}

@article{team2026kimi,
  title={Kimi K2. 5: Visual Agentic Intelligence},
  author={Team, Kimi and Bai, Tongtong and Bai, Yifan and Bao, Yiping and Cai, SH and Cao, Yuan and Charles, Y and Che, HS and Chen, Cheng and Chen, Guanduo and others},
  journal={arXiv preprint arXiv:2602.02276},
  year={2026}
}

@inproceedings{dai2017scannet,
  title={Scannet: Richly-annotated 3d reconstructions of indoor scenes},
  author={Dai, Angela and Chang, Angel X and Savva, Manolis and Halber, Maciej and Funkhouser, Thomas and Nie{\ss}ner, Matthias},
  booktitle={Proceedings of the IEEE conference on computer vision and pattern recognition},
  pages={5828--5839},
  year={2017}
}

@inproceedings{yeshwanth2023scannet++,
  title={Scannet++: A high-fidelity dataset of 3d indoor scenes},
  author={Yeshwanth, Chandan and Liu, Yueh-Cheng and Nie{\ss}ner, Matthias and Dai, Angela},
  booktitle={Proceedings of the IEEE/CVF International Conference on Computer Vision},
  pages={12--22},
  year={2023}
}

@inproceedings{rozenberszki2022language,
  title={Language-grounded indoor 3d semantic segmentation in the wild},
  author={Rozenberszki, David and Litany, Or and Dai, Angela},
  booktitle={European conference on computer vision},
  pages={125--141},
  year={2022},
  organization={Springer}
}

@article{an2025llava,
  title={Llava-onevision-1.5: Fully open framework for democratized multimodal training},
  author={An, Xiang and Xie, Yin and Yang, Kaicheng and Zhang, Wenkang and Zhao, Xiuwei and Cheng, Zheng and Wang, Yirui and Xu, Songcen and Chen, Changrui and Zhu, Didi and others},
  journal={arXiv preprint arXiv:2509.23661},
  year={2025}
}

@article{bai2025qwen3,
  title={Qwen3-vl technical report},
  author={Bai, Shuai and Cai, Yuxuan and Chen, Ruizhe and Chen, Keqin and Chen, Xionghui and Cheng, Zesen and Deng, Lianghao and Ding, Wei and Gao, Chang and Ge, Chunjiang and others},
  journal={arXiv preprint arXiv:2511.21631},
  year={2025}
}

@article{damen2022rescaling,
  title={Rescaling egocentric vision: Collection, pipeline and challenges for epic-kitchens-100},
  author={Damen, Dima and Doughty, Hazel and Farinella, Giovanni Maria and Furnari, Antonino and Kazakos, Evangelos and Ma, Jian and Moltisanti, Davide and Munro, Jonathan and Perrett, Toby and Price, Will and others},
  journal={International Journal of Computer Vision},
  volume={130},
  number={1},
  pages={33--55},
  year={2022},
  publisher={Springer}
}

@article{mangalam2023egoschema,
  title={Egoschema: A diagnostic benchmark for very long-form video language understanding},
  author={Mangalam, Karttikeya and Akshulakov, Raiymbek and Malik, Jitendra},
  journal={Advances in Neural Information Processing Systems},
  volume={36},
  pages={46212--46244},
  year={2023}
}

@article{wu2025spatial,
  title={Spatial-mllm: Boosting mllm capabilities in visual-based spatial intelligence},
  author={Wu, Diankun and Liu, Fangfu and Hung, Yi-Hsin and Duan, Yueqi},
  journal={arXiv preprint arXiv:2505.23747},
  year={2025}
}

@article{cai2025scaling,
  title={Scaling spatial intelligence with multimodal foundation models},
  author={Cai, Zhongang and Wang, Ruisi and Gu, Chenyang and Pu, Fanyi and Xu, Junxiang and Wang, Yubo and Yin, Wanqi and Yang, Zhitao and Wei, Chen and Sun, Qingping and others},
  journal={arXiv preprint arXiv:2511.13719},
  year={2025}
}

@article{yang2025cambrian,
  title={Cambrian-s: Towards spatial supersensing in video},
  author={Yang, Shusheng and Yang, Jihan and Huang, Pinzhi and Brown, Ellis and Yang, Zihao and Yu, Yue and Tong, Shengbang and Zheng, Zihan and Xu, Yifan and Wang, Muhan and others},
  journal={arXiv preprint arXiv:2511.04670},
  year={2025}
}

@article{singh2025openai,
  title={Openai gpt-5 system card},
  author={Singh, Aaditya and Fry, Adam and Perelman, Adam and Tart, Adam and Ganesh, Adi and El-Kishky, Ahmed and McLaughlin, Aidan and Low, Aiden and Ostrow, AJ and Ananthram, Akhila and others},
  journal={arXiv preprint arXiv:2601.03267},
  year={2025}
}

@article{team2023gemini,
  title={Gemini: a family of highly capable multimodal models},
  author={Team, Gemini and Anil, Rohan and Borgeaud, Sebastian and Alayrac, Jean-Baptiste and Yu, Jiahui and Soricut, Radu and Schalkwyk, Johan and Dai, Andrew M and Hauth, Anja and Millican, Katie and others},
  journal={arXiv preprint arXiv:2312.11805},
  year={2023}
}

@inproceedings{yang2025thinking,
  title={Thinking in space: How multimodal large language models see, remember, and recall spaces},
  author={Yang, Jihan and Yang, Shusheng and Gupta, Anjali W and Han, Rilyn and Fei-Fei, Li and Xie, Saining},
  booktitle={Proceedings of the Computer Vision and Pattern Recognition Conference},
  pages={10632--10643},
  year={2025}
}

@inproceedings{li2025sti,
  title={Sti-bench: Are mllms ready for precise spatial-temporal world understanding?},
  author={Li, Yun and Zhang, Yiming and Lin, Tao and Liu, XiangRui and Cai, Wenxiao and Liu, Zheng and Zhao, Bo},
  booktitle={Proceedings of the IEEE/CVF International Conference on Computer Vision},
  pages={5622--5632},
  year={2025}
}

@article{zhu2026video,
  title={Video-MSR: Benchmarking Multi-hop Spatial Reasoning Capabilities of MLLMs},
  author={Zhu, Rui and Shen, Xin and Wu, Shuchen and Miao, Chenxi and Yu, Xin and Li, Yang and Li, Weikang and Xia, Deguo and Huang, Jizhou},
  journal={arXiv preprint arXiv:2601.09430},
  year={2026}
}

@inproceedings{zhang2025open3d,
  title={Open3d-vqa: A benchmark for embodied spatial concept reasoning with multimodal large language model in open space},
  author={Zhang, Weichen and Zhou, Zile and Zeng, Xin and Xuchen, Liu and Fang, Jianjie and Gao, Chen and Cui, Jinqiang and Li, Yong and Chen, Xinlei and Zhang, Xiao-Ping},
  booktitle={Proceedings of the 33rd ACM International Conference on Multimedia},
  pages={12784--12791},
  year={2025}
}

@inproceedings{du2024embspatial,
  title={Embspatial-bench: Benchmarking spatial understanding for embodied tasks with large vision-language models},
  author={Du, Mengfei and Wu, Binhao and Li, Zejun and Huang, Xuan-Jing and Wei, Zhongyu},
  booktitle={Proceedings of the 62nd Annual Meeting of the Association for Computational Linguistics (Volume 2: Short Papers)},
  pages={346--355},
  year={2024}
}

@article{hu2022lora,
  title={Lora: Low-rank adaptation of large language models.},
  author={Hu, Edward J and Shen, Yelong and Wallis, Phillip and Allen-Zhu, Zeyuan and Li, Yuanzhi and Wang, Shean and Wang, Liang and Chen, Weizhu and others},
  journal={Iclr},
  volume={1},
  number={2},
  pages={3},
  year={2022}
}

@article{cheng2025v,
  title={V-star: Benchmarking video-llms on video spatio-temporal reasoning},
  author={Cheng, Zixu and Hu, Jian and Liu, Ziquan and Si, Chenyang and Li, Wei and Gong, Shaogang},
  journal={arXiv preprint arXiv:2503.11495},
  year={2025}
}

@inproceedings{yuan2025videorefer,
  title={Videorefer suite: Advancing spatial-temporal object understanding with video llm},
  author={Yuan, Yuqian and Zhang, Hang and Li, Wentong and Cheng, Zesen and Zhang, Boqiang and Li, Long and Li, Xin and Zhao, Deli and Zhang, Wenqiao and Zhuang, Yueting and others},
  booktitle={Proceedings of the Computer Vision and Pattern Recognition Conference},
  pages={18970--18980},
  year={2025}
}

@inproceedings{
lin2025ostbench,
title={{OST}-Bench: Evaluating the Capabilities of {MLLM}s in Online Spatio-temporal Scene Understanding},
author={Jingli Lin and Chenming Zhu and Runsen Xu and Xiaohan Mao and Xihui Liu and Tai Wang and Jiangmiao Pang},
booktitle={The Thirty-ninth Annual Conference on Neural Information Processing Systems Datasets and Benchmarks Track},
year={2025},
url={https://openreview.net/forum?id=vAkVKIOtcN}
}

@inproceedings{wang2025site,
  title={Site: towards spatial intelligence thorough evaluation},
  author={Wang, Wenqi and Tan, Reuben and Zhu, Pengyue and Yang, Jianwei and Yang, Zhengyuan and Wang, Lijuan and Kolobov, Andrey and Gao, Jianfeng and Gong, Boqing},
  booktitle={Proceedings of the IEEE/CVF International Conference on Computer Vision},
  pages={9058--9069},
  year={2025}
}

@article{huang2024chat,
  title={Chat-scene: Bridging 3d scene and large language models with object identifiers},
  author={Huang, Haifeng and Chen, Yilun and Wang, Zehan and Huang, Rongjie and Xu, Runsen and Wang, Tai and Liu, Luping and Cheng, Xize and Zhao, Yang and Pang, Jiangmiao and others},
  journal={Advances in Neural Information Processing Systems},
  volume={37},
  pages={113991--114017},
  year={2024}
}

@inproceedings{liu2023g,
  title={G-eval: NLG evaluation using gpt-4 with better human alignment},
  author={Liu, Yang and Iter, Dan and Xu, Yichong and Wang, Shuohang and Xu, Ruochen and Zhu, Chenguang},
  booktitle={Proceedings of the 2023 conference on empirical methods in natural language processing},
  pages={2511--2522},
  year={2023}
}

@article{zhang2019bertscore,
  title={Bertscore: Evaluating text generation with bert},
  author={Zhang, Tianyi and Kishore, Varsha and Wu, Felix and Weinberger, Kilian Q and Artzi, Yoav},
  journal={arXiv preprint arXiv:1904.09675},
  year={2019}
}

@inproceedings{papineni2002bleu,
  title={Bleu: a method for automatic evaluation of machine translation},
  author={Papineni, Kishore and Roukos, Salim and Ward, Todd and Zhu, Wei-Jing},
  booktitle={Proceedings of the 40th annual meeting of the Association for Computational Linguistics},
  pages={311--318},
  year={2002}
}

@inproceedings{lin2004rouge,
  title={Rouge: A package for automatic evaluation of summaries},
  author={Lin, Chin-Yew},
  booktitle={Text summarization branches out},
  pages={74--81},
  year={2004}
}

@inproceedings{kato2023arkitscenerefer,
  title={Arkitscenerefer: Text-based localization of small objects in diverse real-world 3d indoor scenes},
  author={Kato, Shunya and Kurita, Shuhei and Chu, Chenhui and Kurohashi, Sadao},
  booktitle={Findings of the Association for Computational Linguistics: EMNLP 2023},
  pages={784--799},
  year={2023}
}

@inproceedings{liang2025referdino, title={Referdino: Referring video object segmentation with visual grounding foundations}, author={Liang, Tianming and Lin, Kun-Yu and Tan, Chaolei and Zhang, Jianguo and Zheng, Wei-Shi and Hu, Jian-Fang}, booktitle={Proceedings of the IEEE/CVF International Conference on Computer Vision}, pages={20009--20019}, year={2025} }

@inproceedings{he2024decoupling, title={Decoupling static and hierarchical motion perception for referring video segmentation}, author={He, Shuting and Ding, Henghui}, booktitle={Proceedings of the IEEE/CVF Conference on Computer Vision and Pattern Recognition}, pages={13332--13341}, year={2024} }

@inproceedings{feng2025object,
  title={Object-shot enhanced grounding network for egocentric video},
  author={Feng, Yisen and Zhang, Haoyu and Liu, Meng and Guan, Weili and Nie, Liqiang},
  booktitle={Proceedings of the Computer Vision and Pattern Recognition Conference},
  pages={24190--24200},
  year={2025}
}

@inproceedings{shen2024learning,
  title={Learning to segment referred objects from narrated egocentric videos},
  author={Shen, Yuhan and Wang, Huiyu and Yang, Xitong and Feiszli, Matt and Elhamifar, Ehsan and Torresani, Lorenzo and Mavroudi, Effrosyni},
  booktitle={Proceedings of the IEEE/CVF Conference on Computer Vision and Pattern Recognition},
  pages={14510--14520},
  year={2024}
}

\clearpage
\appendix
\appendix
\raggedbottom

\begin{center}
{\Large\bfseries Supplementary Material}\\
{\large PinpointQA: A Benchmark for Small Object-Centric
Spatial Understanding in Indoor Videos}
\end{center}

Due to space constraints in the main paper, this appendix supplies
supplementary details and analyses. We begin with additional information
on dataset construction and a more in-depth description of the evaluation
metrics. We then examine representative failure cases and provide
implementation details for the human localization study with FSD guidance.
The appendix concludes with a brief discussion of the benchmark's scope,
limitations, and intended use.

\begin{enumerate}[
    label=\textcolor{navyblue}{\bfseries\Alph*.},
    leftmargin=1.8em,
    itemsep=0.25em
]
    \item \hyperref[sec:A]{\textcolor{navyblue}{\textbf{Details of Dataset Construction}}}
    \begin{enumerate}[
        label=\textcolor{navyblue}{\arabic*.},
        leftmargin=2em,
        itemsep=0.15em
    ]
        \item \hyperref[subsec:A1]{\textcolor{navyblue}{Scene Curation and Target Vocabulary}}
        \item \hyperref[subsec:A2]{\textcolor{navyblue}{Reference Objects and Supporting Surface}}
        \item \hyperref[subsec:A3]{\textcolor{navyblue}{QA Generation}}
        \item \hyperref[subsec:A4]{\textcolor{navyblue}{Quality Control}}
    \end{enumerate}

    \item \hyperref[sec:B]{\textcolor{navyblue}{\textbf{Details of Evaluation Metrics}}}
    \begin{enumerate}[
        label=\textcolor{navyblue}{\arabic*.},
        leftmargin=2em,
        itemsep=0.15em
    ]
        \item \hyperref[subsec:B1]{\textcolor{navyblue}{FSD Evaluation Setting}}
        \item \hyperref[subsec:B2]{\textcolor{navyblue}{SSP Evaluation Setting}}
    \end{enumerate}

    \item \hyperref[sec:C]{\textcolor{navyblue}{\textbf{Representative Qualitative Examples}}}
    \begin{enumerate}[
        label=\textcolor{navyblue}{\arabic*.},
        leftmargin=2em,
        itemsep=0.15em
    ]
        \item \hyperref[subsec:C1]{\textcolor{navyblue}{TPV Failure Case}}
        \item \hyperref[subsec:C2]{\textcolor{navyblue}{NRI Failure Case}}
        \item \hyperref[subsec:C3]{\textcolor{navyblue}{FSD Failure Case}}
        \item \hyperref[subsec:C4]{\textcolor{navyblue}{SSP Failure Case}}
    \end{enumerate}

    \item \hyperref[sec:D]{\textcolor{navyblue}{\textbf{Details of Human Localization with FSD Guidance}}}
    \begin{enumerate}[
        label=\textcolor{navyblue}{\arabic*.},
        leftmargin=2em,
        itemsep=0.15em
    ]
        \item \hyperref[subsec:D1]{\textcolor{navyblue}{Interface and Protocol}}
        \item \hyperref[subsec:D2]{\textcolor{navyblue}{Scoring Scheme}}
    \end{enumerate}

    \item \hyperref[sec:E]{\textcolor{navyblue}{\textbf{Discussion}}}
    \begin{enumerate}[
        label=\textcolor{navyblue}{\arabic*.},
        leftmargin=2em,
        itemsep=0.15em
    ]
        \item \hyperref[subsec:E1]{\textcolor{navyblue}{Intended Use}}
        \item \hyperref[subsec:E2]{\textcolor{navyblue}{Limitations}}
    \end{enumerate}
\end{enumerate}

\vspace{1em}

\section{Details of Dataset Construction}\label{sec:A}

This section supplements the construction pipeline described in the
main paper by providing the complete target vocabulary and concise
details about how the intermediate spatial representations are used
across the four tasks.

\subsection{Scene Curation and Target Vocabulary}\label{subsec:A1}

Scene curation first identifies eligible target objects using a
predefined category vocabulary. The vocabulary contains 109 source
label variants describing portable or movable indoor objects,
particularly objects that can be held in hand or are commonly placed
on tables. After converting each source label to lowercase, we retain
it only when it exactly matches one of these variants.

For benchmark statistics, the 109 source label variants are counted as
102 target categories. We group only source labels that denote the same
reported category. Specifically, \textit{phone}, \textit{smartphone},
and \textit{mobile phone} are grouped as \textit{phone};
\textit{remote control}, \textit{tv remote}, \textit{remote controller},
\textit{remote}, and \textit{tv controller} are grouped as
\textit{remote control}; and \textit{mouse pad} and \textit{mousepad}
are grouped as \textit{mousepad}. These three groups merge ten source
label variants into three reported categories. Each of the remaining
99 source label variants corresponds to a separate category, yielding
102 categories in total. This grouping is used only when reporting
category statistics, while the original source labels are retained in
the generated questions and answers.

\begin{tcolorbox}[
    colback=gray!10,
    colframe=black!80,
    title=Equivalent Labels Used for Category Statistics,
    boxsep=2mm
]
\small
\texttt{phone}, \texttt{smartphone}, \texttt{mobile phone}
$\rightarrow$ \texttt{phone}

\texttt{remote control}, \texttt{tv remote},
\texttt{remote controller}, \texttt{remote},
\texttt{tv controller}
$\rightarrow$ \texttt{remote control}

\texttt{mouse pad}, \texttt{mousepad}
$\rightarrow$ \texttt{mousepad}
\end{tcolorbox}

% \begin{tcolorbox}[
%     colback=gray!10,
%     colframe=black!80,
%     title=Source Label Vocabulary Used for Exact Matching,
%     boxsep=2mm
% ]
\begin{tcolorbox}[
    enhanced,
    breakable,
    colback=gray!10,
    colframe=black!80,
    title=Source Label Vocabulary Used for Exact Matching,
    boxsep=2mm
]
\small
\texttt{bag, backpack, book, bottle, bowl, candle, cup, dumbbell,
hair dryer, hat, headphones, keyboard, laptop, mouse, purse, shoe,
tissue box, water bottle, alarm clock, ball, stuffed animal, plate,
clock, cd case, coffee kettle, paper towel roll, scale, soap dish,
soap dispenser, telephone, toaster, tray, water pitcher,
fire extinguisher, paper cutter, plunger, dustpan, broom, power strip,
toilet paper, guitar, toaster oven, paper bag, key, phone, smartphone,
mobile phone, earbuds, earphones, wireless headphones, headset,
headphone case, remote control, tv remote, remote controller, remote,
tv controller, game controller, charger, power adapter, laptop charger,
phone charger, power bank, wireless charger, power plug, adapter,
dongle, usb hub, extension cord, power cord, cable, wallet, handbag,
tote bag, glasses, eyeglasses, glasses case, mouse pad, mousepad, pen,
pencil, marker, highlighter, notebook, notepad, mug, spray bottle,
lotion bottle, deodorant bottle, toothbrush, razor, comb, hairbrush,
scissors, stapler, tape dispenser, glue bottle, calculator, lanyard,
pen holder, pencil holder, pencil case, post it note, sticky note,
phone stand, tissue, eraser, soap, power extension}
\end{tcolorbox}

\subsection{Reference Objects and Supporting Surface}\label{subsec:A2}

Within each intermediate spatial representation, supporting surfaces
and reference objects serve different roles. The supporting surface is
the primary spatial anchor selected for the target when a suitable
candidate is available, whereas reference objects provide additional
localization cues through their identities, spatial relations, and
distances to the target. NRI excludes the selected supporting surface
and asks for the nearest remaining valid reference. FSD uses the
supporting surface as its main anchor when available and supplements it
with nearby reference objects. SSP preserves the supporting surface
and references as separate fields in the structured output. These
distinct roles allow the same intermediate spatial representation to
support the different localization requirements of the four tasks.

\subsection{QA Generation}\label{subsec:A3}

The four task formats are instantiated independently from the same
intermediate spatial representations. For TPV, positive questions use
retained target labels, while negative questions use vocabulary labels
that are absent from the corresponding scene. For NRI, the correct
answer is the closest valid reference after excluding the supporting
surface, floor and wall categories, and objects in direct contact with
the target. Each NRI question contains three distinct distractors, and
cases with a near tie between the closest reference labels are removed.
For FSD, the selected supporting surface is used as the main anchor
when available; otherwise, the closest valid reference is used. Up to
two additional references are included with their spatial relations and
distances rounded to the nearest centimeter. For SSP, the target, the
selected supporting surface, and up to two closest references are
encoded in a parsable JSON object, with distances retained to one
decimal place. The progression across TPV, NRI, FSD, and SSP therefore
reflects increasingly demanding spatial and output requirements, while
each task is generated directly from the shared intermediate spatial
representation.

\subsection{Quality Control}\label{subsec:A4}

Quality control combines automatic validity checks with iterative
manual review. Automatic checks remove invalid or structural labels,
ambiguous target instances, and QA pairs that do not satisfy the
required output constraints. TPV answers must contain a valid
\textit{Yes} or \textit{No} label. NRI questions must contain four
distinct options and a uniquely determined nearest valid reference.
FSD answers must provide a complete natural language description
grounded in at least one valid spatial anchor. SSP answers must be
directly parsable as JSON objects containing the required
\texttt{target}, \texttt{support\_surface}, and
\texttt{references} fields. Samples or scenes that fail these checks
are excluded.

In each manual review round, we inspect approximately 100 QA pairs
from 10--15 scenes. The review examines whether NRI excludes the
supporting surface from its reference candidates, whether FSD
preserves the intended supporting surface and nearby references,
whether distances and semantic relations remain consistent with the
underlying geometric measurements, whether the question and answer
templates are natural and well formed, and whether the resulting
localization is plausible to a human observer. Problems identified
during a review round are used to revise the QA generation rules,
filtering criteria, and templates before the next round, while the
underlying intermediate spatial representations remain unchanged.

\section{Details of Evaluation Metrics}\label{sec:B}

\subsection{FSD Evaluation Setting}\label{subsec:B1}

FSD predictions are evaluated by GPT-5.4 using a fixed scoring prompt
and a temperature of zero. For each sample, the judge receives the
target label, the ground truth description, and the model prediction.
It evaluates whether the prediction preserves the supporting surface,
reference objects, spatial relations, and numeric distances required
to localize the target. The judge is instructed to assess only
information supported by the ground truth and not to reward additional
references merely because they are plausible in the scene.

The score contains five dimensions with a maximum total of 10 points.
Main location and supporting surface receive up to 3 points, reference
objects receive up to 2 points, spatial relations receive up to
2 points, numeric distances receive up to 2 points, and clarity
receives 1 point. The scoring criteria for these dimensions are
summarized below.

% \begin{tcolorbox}[
%     colback=gray!10,
%     colframe=black!80,
%     title=FSD Scoring Rubric,
%     boxsep=2mm
% ]
\begin{tcolorbox}[
    enhanced,
    breakable,
    colback=gray!10,
    colframe=black!80,
    title=FSD Scoring Rubric,
    boxsep=2mm
]
\small

\textbf{Main location and supporting surface, 0--3 points.}
Three points are assigned when the main location and supporting surface
are correct and the overall placement agrees with the ground truth.
Two points are assigned when the supporting surface or primary anchor
is correct but important localization details are missing. One point
is assigned when the answer identifies only a broad related region
without a reliable supporting surface or precise placement. A clearly
incorrect supporting surface or scene region receives zero points.

\medskip
\textbf{Reference objects, 0--2 points.}
Two points are assigned when at least two correct and relevant
references are used for localization. One point is assigned when
exactly one correct reference is used meaningfully. Unsupported,
incorrect, vague, or merely mentioned references receive no credit.

\medskip
\textbf{Spatial relations, 0--2 points.}
Two points are assigned when the important relations are correct.
One point is assigned when at least one important directional relation
is correct but another is incomplete, less precise, or incorrect.
Predictions in which the important relations are missing, reversed, or
replaced mainly by vague proximity receive no credit. In particular,
replacing a directional relation in the ground truth with
\textit{near} does not constitute an equivalent relation.

\medskip
\textbf{Numeric distances, 0--2 points.}
Two points are assigned when the answer contains usable numeric
distances that are tied to correct references and are roughly
consistent with the ground truth. One point is assigned when at least
one usable distance is correctly grounded but remains incomplete or
moderately inaccurate. A distance receives no credit when it has no
usable unit, is not tied to a correct reference, or is replaced by a
vague expression. The judge uses approximate tolerances of 5 cm for
ground truth distances no greater than 10 cm, 10 cm for distances
between 10 cm and 30 cm, and 15 cm for distances greater than 30 cm.

\medskip
\textbf{Clarity, 0--1 point.}
One point is assigned when the answer is clear and readable. This
dimension evaluates presentation only and cannot compensate for
incorrect spatial content.
\end{tcolorbox}

The judge is required to return a JSON object containing the total
score, the five dimension scores, a concise explanation, and a set of
error tags. The allowed tags are listed below:

\begin{tcolorbox}[
    colback=gray!5,
    colframe=black!50,
    boxsep=1.5mm,
    left=1mm,
    right=1mm,
    top=1mm,
    bottom=1mm
]
\small\raggedright
\path{wrong_main_location},
\path{wrong_support_surface},
\path{missing_key_reference},
\path{wrong_spatial_relation},
\path{missing_numeric_distance},
\path{wrong_numeric_distance},
\path{unclear_expression}, and
\path{hallucinated_reference}.
\end{tcolorbox}

If both the main location score and the reference object score are
zero, the total score cannot exceed 2. Predictions containing almost
no usable localization information are subject to the same upper
bound. After the response is returned, the evaluator checks whether
the total score equals the sum of the five dimension scores and
replaces an inconsistent total with the computed sum. The corrected
total is divided by 10 to obtain the normalized FSD score.

\subsection{SSP Evaluation Setting}\label{subsec:B2}

SSP is evaluated programmatically because its output has a predefined
JSON structure. The evaluator first attempts to parse the complete
model output as a JSON object. If this fails, it removes surrounding
code fences and attempts to recover a JSON object from any additional
text. A prediction receives a score of zero when no JSON object can be
recovered. The recovered object is checked for a
\texttt{support\_surface} field and a \texttt{references} list. The
alternative field name \texttt{primary\_surface} is also interpreted
as \texttt{support\_surface}. Missing or malformed fields receive no
credit for their corresponding components.

Before comparison, object and supporting surface labels are converted
to lowercase, underscores and hyphens are converted to spaces,
unnecessary quotation marks and parenthetical text are removed, and
repeated spaces are collapsed. A conservative singularization rule
and a fixed synonym map are then applied to common equivalent forms.
For example, \textit{mobile phone}, \textit{smartphone}, and
\textit{cell phone} are normalized to \textit{phone};
\textit{night stand} is normalized to \textit{nightstand}; and
\textit{table top} is normalized to \textit{table}. Relation tokens
undergo similar normalization so that forms such as
\textit{beside} and \textit{adjacent to} map to
\texttt{next\_to}, while \textit{close to} and \textit{nearby} map
to \texttt{near}. These operations handle equivalent surface forms
without permitting unrestricted semantic matching.

Since the queried target is already specified in the question, the
target field is not included in the SSP score. The evaluator scores
the supporting surface, reference object identity, spatial relation,
and distance. Supporting surfaces must match exactly after
normalization. Predicted references are aligned one to one with the
ground truth reference slots by selecting the assignment with the
highest agreement in object identity, relation, and distance.
Unmatched ground truth slots receive zero credit.

A reference object receives full credit only when its normalized label
matches the corresponding ground truth label. Relation credit is
computed only after the reference object has been matched. An exact
relation match receives full credit, while the pairs
\texttt{next\_to} and \texttt{near},
\texttt{under} and \texttt{below}, and
\texttt{on} and \texttt{attached\_to}
receive half credit. All other relation mismatches receive zero
credit.

Distance credit requires both a correct reference object and positive
relation credit. Let \(e\) denote the absolute difference between the
predicted and ground truth distances. The exact distance thresholds
are:

\begin{center}
\small
\begin{tabular}{c|c|c}
\hline
\textbf{Ground truth distance} &
\textbf{Full credit} &
\textbf{Half credit} \\
\hline
\(d \leq 10\) cm &
\(e \leq 5\) cm &
\(e \leq 10\) cm \\
\(10 < d \leq 30\) cm &
\(e \leq 10\) cm &
\(e \leq 15\) cm \\
\(d > 30\) cm &
\(e \leq 15\) cm &
\(e \leq 20\) cm \\
\hline
\end{tabular}
\end{center}

Errors beyond the half credit threshold receive zero distance credit.
For each reference slot, the distance credit is multiplied by its
relation credit. Consequently, a half correct relation can contribute
at most half of the available distance credit, even when the numeric
distance itself falls within the full credit threshold. The object,
relation, and distance components are averaged over the ground truth
reference slots.

The final normalized SSP score is computed as
\[
S_{\mathrm{SSP}}
=
0.4S_{\mathrm{surface}}
+
0.2S_{\mathrm{object}}
+
0.2S_{\mathrm{relation}}
+
0.2S_{\mathrm{distance}}.
\]
This procedure separates structural validity from spatial correctness:
a model may produce a parsable JSON object while still receiving a low
score because its supporting surface, references, relations, or
distances are incorrectly grounded.

\section{Representative Qualitative Examples}\label{sec:C}

This section examines representative errors from three scenes with
different target objects: \textit{mouse} in scene
\texttt{062e5a23a6}, \textit{mobile phone} in scene
\texttt{0b031f3119}, and \textit{headphone case} in scene
\texttt{281ba69af1}. Figures~\ref{fig:case_mouse},
\ref{fig:case_phone}, and \ref{fig:case_headphone} show that errors
can occur in different localization components, including target
grounding, nearest reference identification, supporting surface
prediction, reference grounding, spatial relations, numeric distances,
and consistency across structured fields. Green text indicates content
that agrees with the ground truth, while red text indicates incorrect
or unsupported content.

\subsection{TPV Failure Case}
\label{subsec:C1}

Figure~\ref{fig:case_mouse} shows a target grounding error for
\textit{mouse} in scene \texttt{062e5a23a6}. The ground truth confirms
that the mouse appears in the scene, whereas Cambrian-S-7B predicts
\textit{No}. Because TPV requires neither reference selection nor
structured output, this error can be attributed directly to incorrect
target presence verification rather than to a later formatting or
relation error. The example therefore illustrates that a model may
process the broader scene while still failing to identify the queried
small object.

\subsection{NRI Failure Case}
\label{subsec:C2}

The same scene also provides a representative NRI error. As shown in
Figure~\ref{fig:case_mouse}, the correct answer for the nearest valid
reference to the mouse is \textit{office chair}, but GPT-5.4 selects
\textit{laptop bag}. The supporting surface is excluded from the
candidate set in this task, so the error specifically reflects an
incorrect comparison among the remaining reference candidates.
Although the predicted object is present in the scene and appears
plausible as a reference, it is not the closest valid reference defined
by the ground truth.

\subsection{FSD Failure Case}
\label{subsec:C3}

The FSD examples show that a fluent description can still contain
incorrect localization information. In Figure~\ref{fig:case_mouse},
LLaVA-OneVision-1.5-8B describes the \textit{mouse} using a laptop and
keyboard, while the ground truth uses the office chair and glasses case
as the relevant references. The predicted relations and distances are
therefore attached to objects that do not match the ground truth.

Figure~\ref{fig:case_phone} shows a similar error for
\textit{mobile phone}. Kimi K2.5 describes the target relative to a
monitor, power outlets, and keyboard, omitting the tripod that provides
the closest ground truth reference. InternVL3.5-8B-Instruct retains the
table label but replaces the remaining reference configuration with a
keyboard and mouse, together with distances that differ from the
ground truth. In Figure~\ref{fig:case_headphone}, GPT-5.4 places the
\textit{headphone case} on a desk but replaces the speaker and monitor
with a cloth, tissue packet, and plant. LLaVA-OneVision-1.5-8B instead
places the target on the floor and introduces a chair and towel that
are not part of the ground truth localization.

Across these cases, the generated answers remain readable, but their
supporting surfaces, reference objects, relations, or distances are
incorrect. The errors therefore arise from failure to preserve the
grounded spatial information required by FSD rather than from a lack of
language fluency.

\subsection{SSP Failure Case}
\label{subsec:C4}

\begin{figure*}[t]
    \centering
    \includegraphics[
        width=\textwidth,
        trim=8 8 8 8,
        clip
    ]{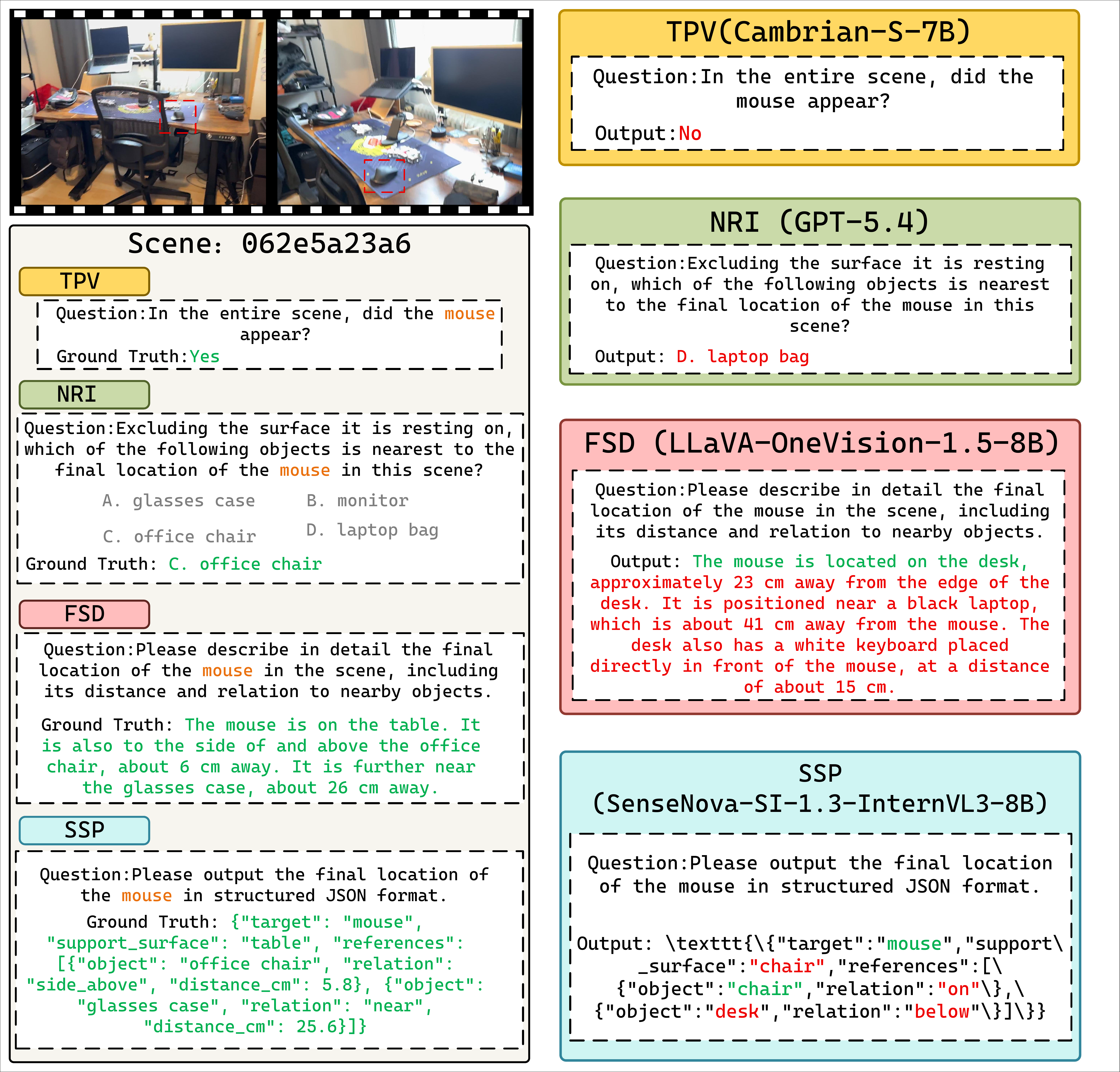}
    \caption{\textbf{Example 1.} Errors across TPV, NRI, FSD, and SSP
    for target \textit{mouse} in scene \texttt{062e5a23a6}. The
    examples show an incorrect target presence decision, an incorrect
    nearest reference, unsupported references and distances in FSD,
    and inconsistent supporting surface, reference, relation, and
    distance fields in SSP. Green text agrees with the ground truth,
    while red text indicates incorrect or unsupported content.}
    \label{fig:case_mouse}
\end{figure*}

The SSP examples demonstrate that producing a parsable JSON object
does not guarantee correct structured spatial prediction. In
Figure~\ref{fig:case_mouse}, SenseNova-SI-1.3-InternVL3-8B predicts
\textit{chair} as the supporting surface instead of \textit{table} and
replaces the ground truth references, \textit{office chair} and
\textit{glasses case}, with \textit{chair} and \textit{desk}. The same
chair is used as both the supporting surface and a reference object,
creating a conflict between the structured fields. Its predicted
relations and distances also differ from the ground truth.

\begin{figure*}[t]
    \centering
    \includegraphics[
        width=\textwidth,
        trim=8 8 8 8,
        clip
    ]{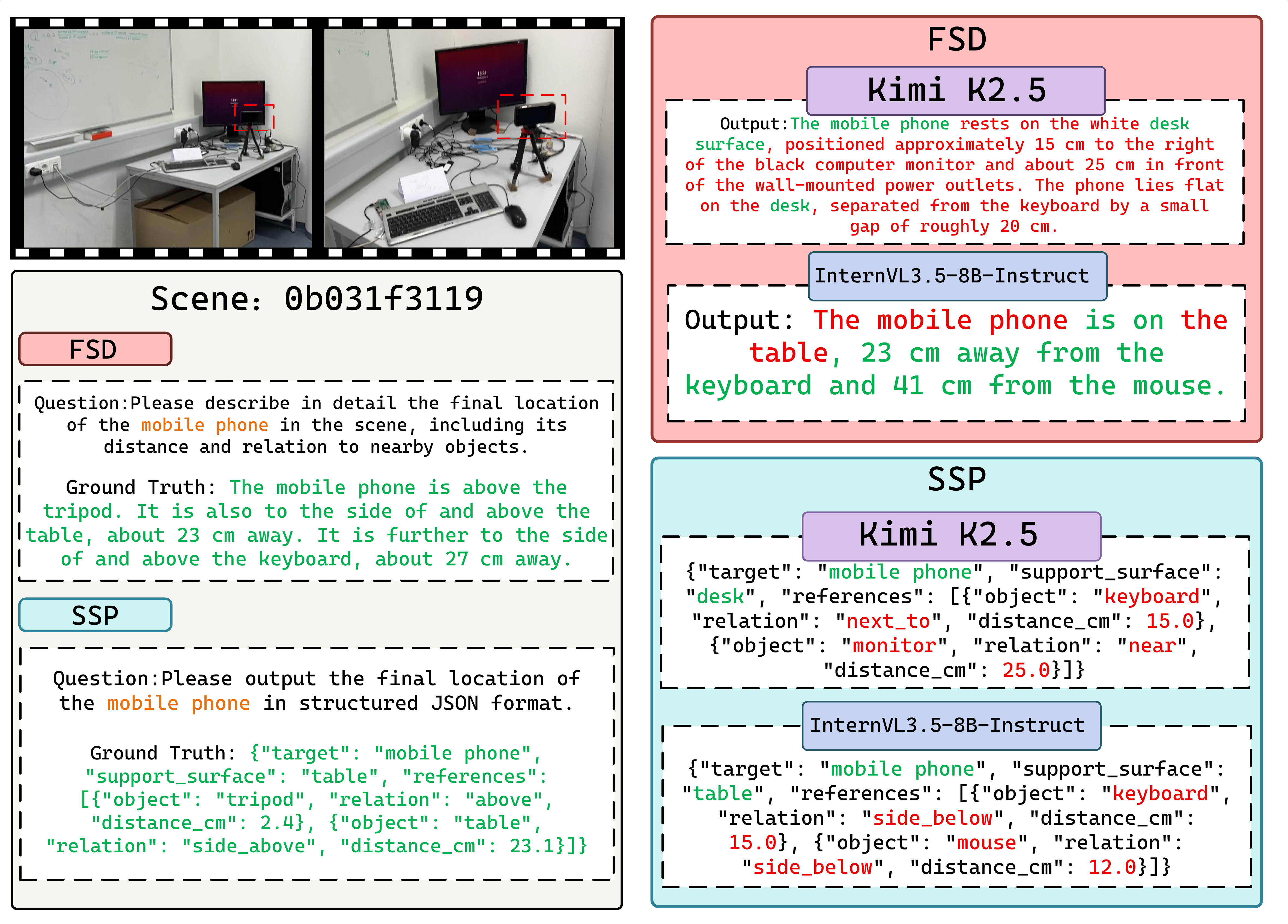}
    \caption{\textbf{Example 2.} FSD and SSP errors for target
    \textit{mobile phone} in scene \texttt{0b031f3119}. Kimi K2.5 and
    InternVL3.5-8B-Instruct produce readable or parsable predictions
    but replace the ground truth references, spatial relations, and
    distances with incorrectly grounded alternatives.}
    \label{fig:case_phone}
\end{figure*}

\begin{figure*}[t]
    \centering
    \includegraphics[
        width=\textwidth,
        trim=8 8 8 8,
        clip
    ]{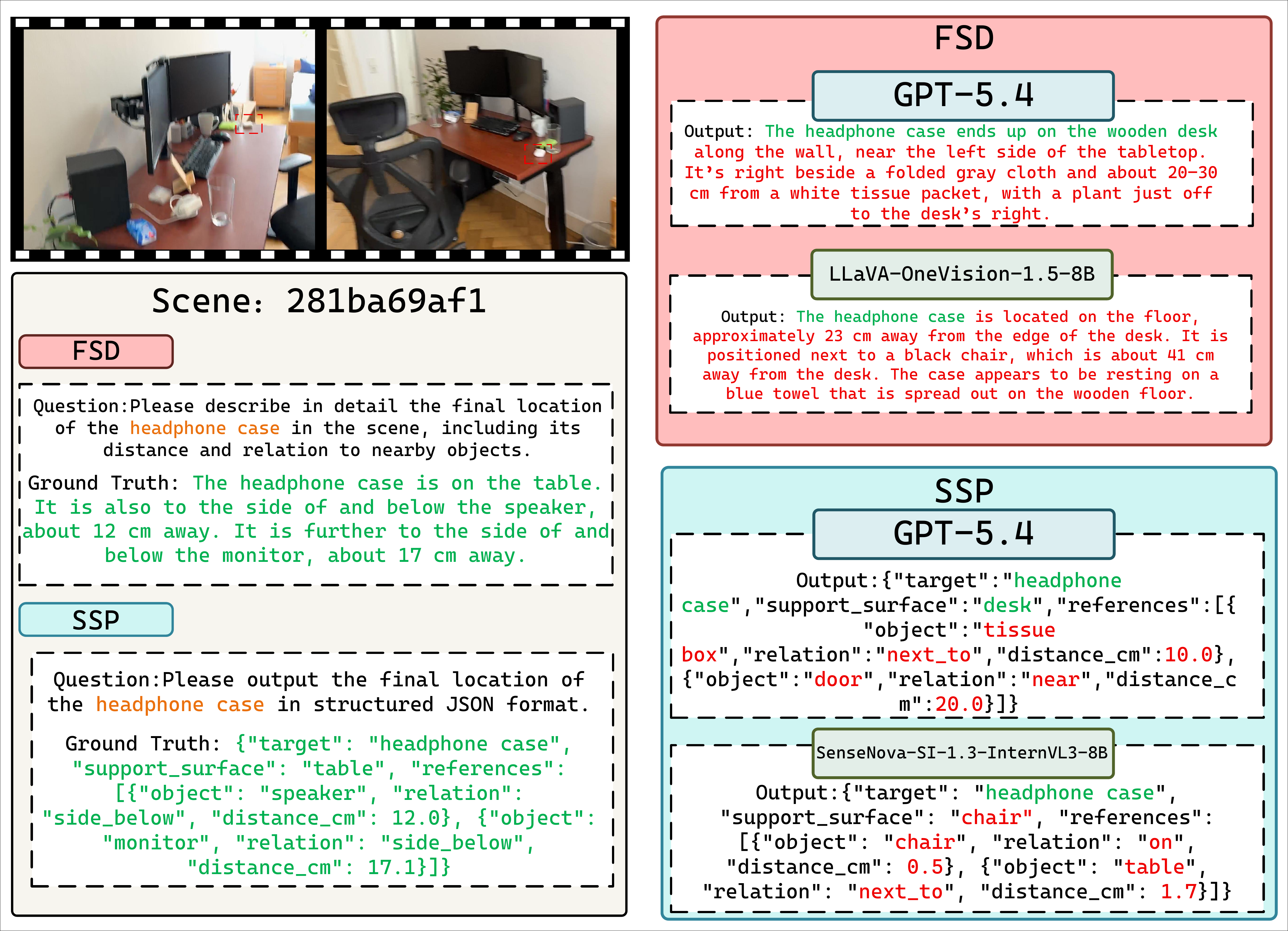}
    \caption{\textbf{Example 3.} FSD and SSP errors for target
    \textit{headphone case} in scene \texttt{281ba69af1}. Although the
    target label is preserved, the predictions contain incorrect
    supporting surfaces, reference objects, spatial relations, or
    distances.}
    \label{fig:case_headphone}
\end{figure*}

Figure~\ref{fig:case_phone} presents two additional structured
prediction errors. Kimi K2.5 produces valid JSON but changes the
supporting surface from \textit{table} to \textit{desk} and replaces
the ground truth references with \textit{keyboard} and
\textit{monitor}. InternVL3.5-8B-Instruct preserves the table as the
supporting surface but predicts \textit{keyboard} and \textit{mouse}
as references, with incorrect relations and distances. Similarly, in
Figure~\ref{fig:case_headphone}, GPT-5.4 predicts \textit{tissue box}
and \textit{door} instead of \textit{speaker} and \textit{monitor},
while SenseNova-SI-1.3-InternVL3-8B changes the supporting surface to
\textit{chair} and introduces a different reference configuration.

These cases cover supporting surface mismatches, reference
hallucination or drift, relation inconsistencies, inaccurate
distances, and conflicts across structured fields. They also show why
low SSP performance cannot be explained only by failures to generate
valid JSON: an output may satisfy the requested structure while its
localization fields remain inconsistent with the ground truth.

\section{Details of Human Localization with FSD Guidance}
\label{sec:D}

\subsection{Interface and Protocol}
\label{subsec:D1}

\begin{figure*}[t]
    \centering
    \includegraphics[width=\textwidth]{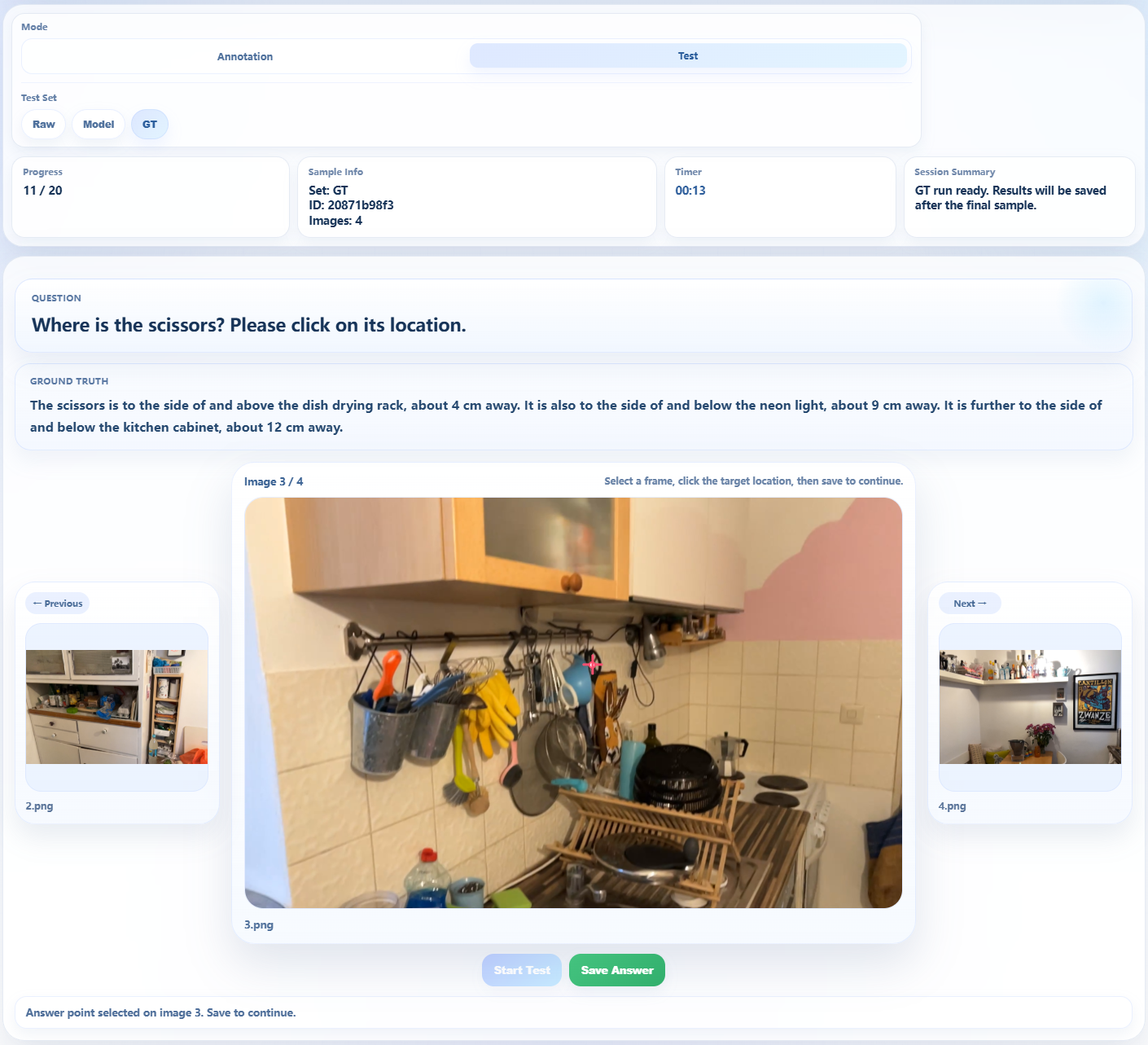}
    \caption{Human evaluation interface in the GT-Assisted setting.}
    \label{fig:human_interface}
\end{figure*}

We implement the human localization study described in Section~4.4 of the main paper using a web interface. For each
sample, the interface presents the question together with the sampled
video frames. Participants browse the frames, select the frame in which
they believe the target appears, click its estimated location, and save
their response. The visual interaction remains the same across the three
settings, while the available textual guidance differs. The Unguided
setting contains no spatial description. The Model-Assisted setting
displays the FSD generated by Qwen3-VL-8B-Instruct-SFT, whereas the
GT-Assisted setting displays the corresponding ground truth FSD.
Figure~\ref{fig:human_interface} shows the interface in the GT-Assisted
setting.

Timing begins when the participant starts the trial and the video frames
are revealed, and ends when the participant saves the response. The
recorded completion time therefore includes both frame selection and
point localization.
Before the study, we annotate the target location using the same interface
in annotation mode. For each sample, this process records the reference
frame index and the target coordinates normalized by the image dimensions.

\subsection{Scoring Scheme}
\label{subsec:D2}

Each submitted response consists of a selected frame index and normalized
click coordinates $(x,y)$. Let $(x^{*},y^{*})$ denote the annotated target
coordinates. When the selected frame is correct, the normalized Euclidean
distance between the submitted and annotated points is computed as
\[
d =
\sqrt{
    (x-x^{*})^{2}
    +
    (y-y^{*})^{2}
}.
\]

An incorrect frame receives a score of $0$. For a correct frame, we use
a fixed radius of $r=0.12$ and compute the score as
\[
s =
\max\left(
0,\,
1-\left(\frac{d}{r}\right)^2
\right).
\]
The score decreases quadratically with the distance from the annotated
target and becomes zero when the distance reaches $r$.

For $N$ evaluated responses, the reported localization accuracy is
\[
\mathrm{Accuracy}
=
\frac{100}{N}
\sum_{i=1}^{N}s_i.
\]
The reported completion time is the mean elapsed time across the same
responses. This protocol jointly evaluates correct frame selection and
accurate target localization. As stated in the main paper, the study
provides auxiliary evidence about the usefulness of FSD guidance for
human localization rather than a deployment evaluation.

\section{Discussion}
\label{sec:E}

\subsection{Intended Use}
\label{subsec:E1}

PinpointQA is intended as a focused diagnostic benchmark and supervision
resource for small object-centric spatial understanding in indoor videos.
It does not aim to cover general video understanding or provide a corpus
for large scale web pretraining. Instead, its four tasks examine a
progressive capability chain from target presence verification to
reference grounding, fine-grained spatial description, and structured
spatial prediction. This design supports analysis of whether a model can
identify a small target and accurately recover its supporting surface,
nearby reference objects, spatial relations, and distances from video
observations.

In addition to evaluation, the benchmark can provide supervision for the
spatial QA tasks defined in this work. The improvements obtained by the
two models after supervised fine-tuning demonstrate this use within the
scope of PinpointQA. These results should not be interpreted as evidence
of broader improvements in general video understanding or unconstrained
embodied reasoning.

\subsection{Limitations}
\label{subsec:E2}

PinpointQA is constructed from ScanNet++ and ScanNet200 and is restricted
to real indoor scenes and the selected vocabulary of small object
categories. It therefore does not establish generalization to outdoor
environments, other data sources, or unconstrained object vocabularies.
Evaluating generalization across datasets and environments remains an
important direction for future work.

The benchmark focuses on spatial grounding from video observations rather
than general temporal understanding, open-world object tracking, or
complete scene interpretation. Its four tasks end with natural language
and structured spatial prediction and do not directly evaluate downstream
navigation, manipulation, or physical interaction. PinpointQA should
therefore be viewed as evaluating a perceptual and spatial capability that
precedes these embodied behaviors rather than the behaviors themselves.

The QA pairs are instantiated from intermediate spatial representations
using controlled construction rules. This design provides consistent and
reproducible supervision across the four tasks, but it also produces more
regular question and answer patterns than unconstrained human language.
The benchmark consequently prioritizes reliable spatial grounding and
comparability over linguistic diversity. Finally, the released benchmark
does not redistribute the original videos or raw scene assets, so complete
reproduction requires access to the corresponding source datasets.

\end{document}